\documentclass[twoside,leqno,twocolumn]{article}

\usepackage[letterpaper]{geometry}

\usepackage{ltexpprt}
\usepackage{hyperref}

\usepackage{cite}
\usepackage{amsmath,amssymb,amsfonts}
\usepackage{algorithmic}
\usepackage{graphicx}
\usepackage{textcomp}
\usepackage{xcolor}
\usepackage{multirow}

\usepackage{booktabs}
\usepackage{algorithm}
\usepackage{mathrsfs}
\usepackage{makecell}
\usepackage{enumitem}

\usepackage[T1]{fontenc}% http://ctan.org/pkg/fontenc
\usepackage[outline]{contour}% http://ctan.org/pkg/contour
\contourlength{0.1pt}
\contournumber{10}%

\DeclareMathSymbol{:}{\mathord}{operators}{"3A}
\usepackage{tabularx}

\begin{document}

\newcommand\relatedversion{}
\renewcommand\relatedversion{\thanks{The full version of the paper can be accessed at \protect\url{https://arxiv.org/abs/1902.09310}}} % Replace URL with link to full paper or comment out this line

%\setcounter{chapter}{2} % If you are doing your chapter as chapter one,
%\setcounter{section}{3} % comment these two lines out.

% \title{\Large Semi-Supervised Clustering via Structural Entropy \\ with Different Types of Constraints}
\title{\Large Semi-Supervised Clustering via Structural Entropy with Different Constraints}
% \author{Corey Gray\thanks{Society for Industrial and Applied Mathematics.}
% \and Tricia Manning\thanks{Society for Industrial and Applied Mathematics.}}
% \author{Anonymous}
% \author{Guangjie Zeng\thanks{Beihang University.State Key Laboratory of Software Development Environment}
% \and Hao Peng\thanks{Beihang University.}
% \and Angsheng Li\thanks{Beihang University.}
% \and Zhiwei Liu\thanks{Salesforce AI Research.}
% \and Chunyang Liu\thanks{Didi Chuxing.}
% \and Lifang He\thanks{Lehigh University.}}
\author{Guangjie Zeng\thanks{SKLSDE, Beihang University. \{zengguangjie, penghao, angsheng, yangrunze\}@buaa.edu.cn}
\and Hao Peng\footnotemark[1]
\and Angsheng Li\footnotemark[1]
\and Zhiwei Liu\thanks{Salesforce AI Research. zhiweiliu@salesforce.com} 
\and Runze Yang\footnotemark[1]
\and Chunyang Liu\thanks{Didi Chuxing. liuchunyang@didiglobal.com}
\and Lifang He\thanks{Lehigh University. lih319@lehigh.edu}}

\date{}

\maketitle

% Copyright Statement
% When submitting your final paper to a SIAM proceedings, it is requested that you include
% the appropriate copyright in the footer of the paper.  The copyright added should be
% consistent with the copyright selected on the copyright form submitted with the paper.
% Please note that "20XX" should be changed to the year of the meeting.

% Default Copyright Statement
\fancyfoot[R]{\scriptsize{Copyright \textcopyright\ 2024 by SIAM\\
Unauthorized reproduction of this article is prohibited}}

% Depending on which copyright you agree to when you sign the copyright form, the copyright
% can be changed to one of the following after commenting out the default copyright statement
% above.

%\fancyfoot[R]{\scriptsize{Copyright \textcopyright\ 20XX\\
%Copyright for this paper is retained by authors}}

%\fancyfoot[R]{\scriptsize{Copyright \textcopyright\ 20XX\\
%Copyright retained by principal author's organization}}

%\pagenumbering{arabic}
%\setcounter{page}{1}%Leave this line commented out.

\begin{abstract} \small\baselineskip=9pt 
Semi-supervised clustering techniques have emerged as valuable tools for leveraging prior information in the form of constraints to improve the quality of clustering outcomes.
% While many semi-supervised clustering methods have been proposed, few can incorporate different types of constraints.
Despite the proliferation of such methods, the ability to seamlessly integrate various types of constraints remains limited.
% In the meantime, structural entropy although a powerful clustering method that has achieved great success in many research areas, lacks a variant that can incorporate these constraints.
While structural entropy has proven to be a powerful clustering approach with wide-ranging applications, it has lacked a variant capable of accommodating these constraints.
In this work, we present \underline{S}emi-supervised clustering via \underline{S}tructural \underline{E}ntropy (SSE), a novel method that can incorporate different types of constraints from diverse sources to perform both partitioning and hierarchical clustering.
Specifically, we formulate a uniform view for the commonly used pairwise and label constraints for both types of clustering.
Then, we design objectives that incorporate these constraints into structural entropy and develop tailored algorithms for their optimization.
We evaluate SSE on nine clustering datasets and compare it with eleven semi-supervised partitioning and hierarchical clustering methods.
Experimental results demonstrate the superiority of SSE on clustering accuracy with different types of constraints.
Additionally, the functionality of SSE for biological data analysis is demonstrated by cell clustering experiments conducted on four single-cell RNA-seq datasets.
\end{abstract}

\noindent\textbf{Keywords:} Semi-Supervised Clustering, Structural Entropy, Biological Data Analysis.

% 1. Abstract 第5局，直接告诉读者你做了什么，不要说你分析了多少多少，分析不是一个创造性的动词；

% 2. Intro第一段需要增加代表性的引用;

% 3. 优势什么什么？ 聚类精度，还是聚类计算效率，你要clear说出来几个优势；

% 4. Intro里面要加一个toy example，让读者看到你做了啥；

% 基于结构信息理论的半监督图聚类with.......

\section{Introduction}
Clustering is a key technique in machine learning that aims to group instances according to their similarity \cite{gan2020data}.
Yet, unsupervised clustering alone often fails to provide the desired level of accuracy and may not meet the diverse requirements of various users.
In contrast, semi-supervised clustering harnesses the power of prior information in the form of constraints, significantly boosting clustering accuracy and aligning more effectively with user preferences \cite{nie2021semi}.
% In contrast, semi-supervised clustering leverages prior information in the form of constraints to enhance clustering accuracy and align with user preferences \cite{nie2021semi}.

Numerous semi-supervised clustering methods based on different classical unsupervised clustering methods have been proposed in recent years.
% The challenge of semi-supervised clustering is to design an objective that incorporates these constraints into classical clustering methods reasonably and to find an optimization algorithm that optimizes the objective effectively and efficiently.
The challenges of semi-supervised clustering are 1) to design an objective function integrating constraints into clustering methods and 2) to effectively and efficiently optimize the objective.
% The most used way to utilize the prior information is to add a penalty term on the original clustering objective \cite{nie2021semi,jiang2022semi}, \cite{lipor2017leveraging,lan2020label}.
% aim to learn an adjusted distance measure from given prior information.
% The most widely-used way to utilize the prior information is to add a penalty term on the original clustering objective \cite{nie2021semi,jiang2022semi}, while some others propagate them to enhance the data \cite{lipor2017leveraging,lan2020label}.
The most widely-used way to utilize the prior information is to add a regularization on the original clustering objective \cite{nie2021semi,jiang2022semi}.
Alternatively, some methods propagate this information to augment the dataset itself \cite{lipor2017leveraging,lan2020label}.
The provided prior information can manifest in various constraint forms, such as pairwise constraints \cite{wagstaff2001constrained}, and label constraints \cite{liu2023constrained}, and triplet constraints \cite{zheng2011semi}.
Many existing semi-supervised clustering methods are tailored to handle a single type of constraint.
% However, the given prior information can be in different forms from multiple sources. 
Yet, it is common for prior information to come in diverse forms from multiple sources.
The lack of ability to deal with different types of constraints limits the generalization ability of these methods.
% present a figure here to illustrate this problem.

% In order to incorporate different types of constraints into the semi-supervised clustering methods, earlier methods \cite{davidson2005agglomerative,xiao2016semi} discuss them case by case with different algorithms.
Concerning the integration of different types of constraints into the semi-supervised clustering methods, earlier methods 
\cite{davidson2005agglomerative,xiao2016semi} discuss them case by case with different algorithms.
However, these methods lack a unified view of constraints and are unable to deal with mixed types of constraints.
% APJCSC \cite{qian2016affinity} transforms group constraints into pairwise constraints to incorporate multiple types of constraints, but the explicit transformation is not formally formulated the unified view of constraints is not given.
% Bai \textit{et al.} gave a uniform representation of pairwise constraints and label constraints \cite{bai2020semi} and proposed the SC-MPI algorithm based on the spectral clustering algorithm.
Bai \textit{et al.} resolved this issue via a unified formulation of pairwise constraints and label constraints \cite{bai2020semi} and proposed the SC-MPI algorithm to optimize them simultaneously.
% Both pairwise constraints and label constraints are stored in a relation matrix and are used as a penalty term in the objective of SC-MPI optimized by eigenvalue decomposition.
% However, SC-MPI requires a prior known parameter $k$ as the number of clusters, which is not accessible in many real scenarios.
% However, SC-MPI lacks the generalization ability for hierarchical clustering, which is important for many applications such as cell subtype finding in biological data analysis.
However, SC-MPI, which is designed for partitioning clustering, cannot perform hierarchical clustering and thus has limited generalization ability.
Hierarchical clustering does not require specifying the number of clusters in advance, and it produces a dendrogram that shows the nested structure of the data. 
This is useful for many applications, such as finding cell subtypes in biological data analysis \cite{10.1093/nar/gkac1044}.
% SC-MPI incorporates these constraints into the spectral clustering algorithm as a penalty term and optimizes the new objective with the eigenvalue decomposition method.
% However, eigenvalue decomposition-based optimization of SC-MPI is time-consuming and limits its application to large-scale datasets, and SC-MPI requires a prior known $k$ to as the cluster number.
% Moreover, other types of constraints such as triplet constraints can not be incorporated into SC-MPI, which limits its generalization ability.

% \begin{figure}[t]
%     \centering
%     \includegraphics[width=1\linewidth]{figures/challenge.pdf}
%     \vspace{-0.5cm}
%     \caption{Comparison of structural entropy-based clustering and our proposed SSE clustering, on dataset.}
%     \vspace{-0.5cm}
%     \label{fig:1}
% \end{figure}

% In this work, we propose a \underline{S}emi-supervised clustering method via \underline{S}tructural \underline{E}ntropy with different types of constraints, namely SSE, as a more general semi-supervised clustering method for both partitional clustering and hierarchical clustering.
% In order to address above issues, we propose a \underline{S}emi-supervised clustering method via \underline{S}tructural \underline{E}ntropy with different constraints, namely SSE, as a more general semi-supervised clustering method for both partitional clustering and hierarchical clustering.
To address aforementioned issues, we propose a more general \underline{S}emi-supervised clustering method via \underline{S}tructural \underline{E}ntropy with different constraints, namely SSE, for both partitioning clustering and hierarchical clustering.
% The basic idea is that we introduce a penalty term that changes the cut of the original structural entropy objective, penalizing clusters that violate must-link constraints and rewarding clusters that obey cannot-link constraints.
% First, we construct a graph $G$ and a relation graph $G'$ to represent the information of data and prior information in constraints respectively.
First, we construct a data graph $G$ and a relation graph $G'$ sharing the same set of vertices to represent the information of input data and prior information in constraints, respectively.
% The vertices in $G$ are instances of data and edge weights are the similarity between instances.
% Data points are vertices in $G$ and the similarities between them are edge weights.
Vertices and edge weights of $G$ are data points and similarities between them, respectively.
% Different types of constraints are transformed into a uniform view and stored in $G'$ with edge weights representing the relationship between data points.
Different types of constraints are formulated as a uniform view and stored in $G'$ with positive edge weights representing must-link relationships and negative weights representing cannot-link relationships between data points.
% Second, we devise the objective of two-dimensional SSE for semi-supervised partitional clustering by adding a penalty term on the objective of two-dimensional structural entropy and then optimize it through two operators \textit{merging} and \textit{moving}.
Second, we devise the objective of two-dimensional (2-d) SSE for semi-supervised partitioning clustering by adding a penalty term to the objective of 2-d structural entropy and then optimize it through two modified operators \textit{merging} and \textit{moving}.
Third, we devise the objective of high-dimensional (high-d) SSE for semi-supervised hierarchical clustering by extending the objective of 2-d SSE, and then optimize it through two modified operators \textit{stretching} and \textit{compressing}.
A binary encoding tree is obtained by \textit{stretching} and an encoding tree with a certain height is obtained by \textit{compressing}.
The source code is available on GitHub\footnote{https://github.com/SELGroup/SSE}.

% We comprehensively evaluate SSE with semi-supervised clustering methods with two types of constraints.
We comprehensively evaluate SSE regarding semi-supervised clustering methods with respect to two types of constraints.
% The results show that SSE achieves better performance with both types of constraints.
The results justify the better performance of SSE under both types of constraints.
We also conduct experiments on four single-cell RNA-seq datasets to perform cell clustering, demonstrating the functionality of SSE for biological data analysis.
The main contributions of this paper are summarized as follows:
% (1) We give a uniform representation form of three types of constraints including pairwise constraints, group constraints, and triplet constraints, and formulate them into a penalty term added to the structural entropy objective to form the objective of SSE.
% \begin{itemize}
%     \item We give a uniform representation of pairwise constraints and label constraints and use them in a penalty term to form the objective of SSE.
%     \item We design efficient algorithms to optimize the objective of SSE to enable semi-supervised partitional clustering and hierarchical clustering.
%     \item SSE achieves the best performance compared to both semi-supervised partitional clustering and hierarchical clustering methods.
% \end{itemize}
% (1) We give a uniform representation of pairwise constraints and label constraints and use them in a penalty term to form the objective of SSE.
(1) We devise a uniform formulation for pairwise constraints and label constraints and use them in a penalty term to form the objective of SSE.
(2) We design efficient algorithms to optimize the objective of SSE to enable semi-supervised partitioning clustering and hierarchical clustering.
% (3) SSE achieves the best performance compared to both semi-supervised partitional clustering and hierarchical clustering methods.
(3) The extensive experiments on nine clustering datasets and four single-cell RNA-seq datasets indicate that SSE achieves the best performance among semi-supervised clustering methods and is effective for biological data analysis.
\section{Structural Entropy}
% We briefly summarize structural entropy \cite{li2016structural} in this section.
We provide a brief introduction to structural entropy \cite{li2016structural} before presenting our model.
% Structural entropy is based on the idea of encoding trees to capture the uncertainty of graphs' hierarchical structure, which is naturally a way of vertices clustering.
Intuitively, structural entropy methods encode tree structures via characterizing the uncertainty of the hierarchical topology.
% The structural entropy of a graph $G$ is defined as the minimum total number of bits required to determine the codewords of the nodes in $G$.
The structural entropy of a graph $G$ is defined as the minimum total number of bits required to determine the codewords of nodes in $G$.
% Structural entropy makes great success in the field of bioinformatics \cite{10.1093/nar/gkac1044}, Reinforcement Learning \cite{zeng2023effective}, and graph neural networks \cite{wu2022structural}.
Structural entropy has achieved success in the field of traffic forecast \cite{zou2023multispans}, social event detection \cite{cao2024hierarchical}, and reinforcement learning \cite{zeng2023effective,zeng2024adversarial}.
% Through minimizing the structural entropy of a given graph $G$, the hierarchical clustering result of vertices in $G$ is represented by the associated encoding tree.
Through minimizing the structural entropy of a given graph $G$, the hierarchical clustering result of vertices in $G$ is retained by the associated encoding tree.

\noindent\textbf{Encoding tree.}
Let $G=(V,E,\textbf{W})$ be an undirected weighted graph, where $V=\{v_1,...,v_n\}$ is the vertex set, $E$ is the edge set, and $\textbf{W} \in \mathbb{R}^{n \times n}$ is the edge weight matrix.
The encoding tree $\mathcal{T}$ of $G$ as a hierarchical rooted tree is defined as follows:
(1) For each tree node $\alpha \in \mathcal{T}$, a vertex subset $T_\alpha \in V$ is associated with it.
(2) The root node $\lambda$ of $\mathcal{T}$ is associated with the vertex set $V$, i.e., $T_{\lambda} = V$.
(3) For each $\alpha \in \mathcal{T}$, the immediate successors of it are labeled by $\alpha^{\wedge}\langle i\rangle$ ordered from left to right as $i$ increases, and the immediate predecessor of it is written as $\alpha^-$.
(4) For each $\alpha \in \mathcal{T}$ with $L$ immediate successors, vertex subsets $T_{\alpha^{\wedge}\langle i\rangle}$ are disjoint and $T_{\alpha}= \cup_{i=1}^{L} T_{\alpha^{\wedge}\langle i\rangle}$.
(5) For each leaf node $\nu \in \mathcal{T}$, $T_\nu$ contains only one vertex in $V$.

\noindent\textbf{$K$-D Structural Entropy.}
Given an arbitrary rooted encoding tree $\mathcal{T}$ of a graph $G$, the structural entropy of $G$ on $\mathcal{T}$ measures the amount of remaining complexity in $G$ after reduced by $\mathcal{T}$.
For each non-root node $\alpha \in \mathcal{T}$, the assigned structural entropy of it is defined as:
\begin{equation}\label{eq:ht}
    \mathcal{H}^\mathcal{T}(G;\alpha) = - \frac{g_\alpha}{\mathcal{V}_G}\log_2\frac{\mathcal{V}_\alpha}{\mathcal{V}_{\alpha^-}},
\end{equation}
where $g_\alpha$ is the cut, i.e., the weight sum of edges between nodes in and not in $T_\alpha$, $\mathcal{V}_\alpha$ and $\mathcal{V}_G$ are the volumes, i.e., the sum of node degrees in $T_\alpha$ and $G$, respectively.
The structural entropy of $G$ given by $\mathcal{T}$ is defined as:
\begin{equation}
    \mathcal{H}^\mathcal{T}(G) = \sum_{\alpha\in\mathcal{T},\alpha\neq\lambda}\mathcal{H}^\mathcal{T}(G;\alpha).
\end{equation}
To meet the requirements of downstream applications, the $K$-dimensional structural entropy of $G$ is defined as:
\begin{equation}
    \mathcal{H}^K(G) = \min_\mathcal{T}\{\mathcal{H}^\mathcal{T}(G)\},
\end{equation}
where $\mathcal{T}$ ranges over all encoding trees whose heights are at most $K$.
% When the height of the encoding tree is limited to at most two, the encoding tree of $G$ gives a partitioning clustering result of graph vertices; when the height of the encoding tree is larger than two, the encoding tree of $G$ gives a hierarchical clustering result of graph vertices.

\noindent\textbf{2-D Structural Entropy}.
One special case of $K$-d structural entropy is 2-d structural entropy, where the encoding tree represents a graph partitioning.
% Given an undirected weighted graph $G=(V,E,\mathbf{W})$, the two-dimensional structural entropy of $G$ is the minimum structural entropy on all possible encoding trees with heights at most two.
A 2-d encoding tree $\mathcal{T}$ can be formulated as a graph partitioning $\mathcal{P}=\{X_1, X_2,..., X_L\}$ of $V$, where $X_i$ is a vertex subset called module associated with the $i$-th children of root $\lambda$.
The structural entropy of $G$ given by $\mathcal{P}$ is defined as:
\begin{equation}
\label{eq:hp}
\begin{aligned}
    \mathcal{H}^\mathcal{P}(G) = - \sum_{X \in \mathcal{P}}\sum_{v_i \in X} \frac{g_i}{\mathcal{V}_G} \log_2 \frac{d_i}{\mathcal{V}_X} \\
    - \sum_{X \in \mathcal{P}} \frac{g_X}{\mathcal{V}_G} \log_2 \frac{\mathcal{V}_X}{\mathcal{V}_G},
\end{aligned}
\end{equation}
where $d_i$ is the degree of vertex $v_i$, $g_i$ is the cut, i.e., the weight sum of edges connecting $v_i$ and other vertices, $\mathcal{V}_X$ and $\mathcal{V}_G$ are the volumes, i.e., the sum of node degrees in module $X$ and graph $G$, respectively, and $g_X$ is the cut, i.e., the weight sum of edges between vertices in and not in module $X$.
\section{Methodology}
\begin{figure*}[t]
\centering
\includegraphics[width=1\linewidth]{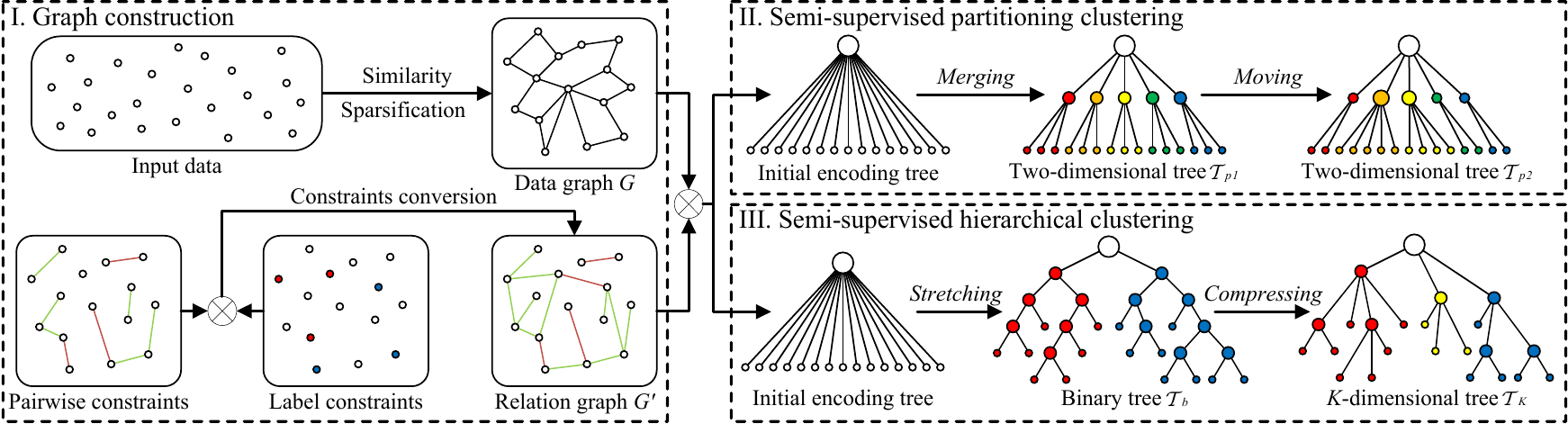}
\vspace{-0.6cm}
% \captionsetup{width=0.8\textwidth}
\caption{Overview of SSE. (I) Two graphs $G$ and $G'$ are constructed from input data and constraints, respectively. (II) Semi-supervised partitioning clustering is performed through two operators \textit{merging} and \textit{moving}. (III) Semi-supervised hierarchical clustering is performed through two operators \textit{stretching} and \textit{compressing}.}
\label{fig:framework}
\end{figure*}

In this section, we present the proposed SSE algorithm for semi-supervised clustering.
% The framework of SSE is depicted in Fig. \ref{fig:framework}.
The framework of SSE is depicted in Figure \ref{fig:framework}.
% SSE is composed of three components: graph construction, semi-supervised partitional clustering, and semi-supervised hierarchical clustering.
SSE has three components: graph construction, semi-supervised partitioning clustering, and semi-supervised hierarchical clustering.
% Data points and constraints are transformed into graphs, and are then used to perform semi-supervised partitional clustering and semi-supervised hierarchical clustering through two-dimensional SSE and high-dimensional SSE minimization respectively.
Input data and constraints are transformed into two different graphs sharing the same vertex set and then used to perform semi-supervised partitioning clustering and semi-supervised hierarchical clustering through 2-d SSE and high-d SSE minimization, respectively.

% \subsection{Uniform Representation of Constraints.}\label{Sec:constraints}
\subsection{Uniform Formulation of Constraints.}\label{sec:constraints}
Considering a graph $G=(V,E,\mathbf{W})$ associated to a given dataset $\mathcal{X}=\{x_1,x_2,...,x_n\}$, where $x_i$ is a data point, $V=\{v_1,v_2,...,v_n\}$ correspond to data points in $\mathcal{X}$, the edges in $E$ connect similar data points, and edge weights in $\mathbf{W}$ represent similarity of data points.
We aim to partition graph vertices in $G$ with certain given prior information in the form of constraints to achieve semi-supervised data clustering.
The pairwise constraints and label constraints are formulated as follows.

% Pairwise constraints include must-link constraints and cannot-link constraints, specifying a pair of data points must belong to the same and different clusters respectively. 
Pairwise constraints reveal the relationship between a pair of vertices in $G$.
% They can be formulated as a set of must-link constraints $\mathcal{M}=\{(x_i,x_j):x_i \in \}$
They consist of a set of must-link constraints $M=\{(v_i,v_j): \: l_i=l_j\}$, indicating that vertex pair $(v_i,v_j)$ must belong to the same cluster, and a set of cannot-link constraints $C=\{(x_i,x_j): \: l_i\neq l_j\}$, indicating that vertex pair $(v_i,v_j)$ must belong to different clusters, where $l_i$ is the cluster indicator of $v_i$.
Pairwise constraints can be stored in a relation graph $G'=(V, E',\mathbf{W'})$, which shares the same vertex set with $G$.
If there exists a vertex pair $(v_i, v_j) \in M$, an edge exists in $E'$ with a positive value $\gamma_M$ added to the edge weight $\mathbf{W}'_{ij}$.
If there exists a vertex pair $(v_i, v_j) \in C$, an edge exists in $E'$ with a negative value $\gamma_C$ added to the edge weight $\mathbf{W}'_{ij}$.
The values of $\mathbf{W}$ and $\mathbf{W'}$ are set in \textbf{Implementation Details} in Section \ref{sec:experiments}.

Label constraints reveal the relationship between vertices in $G$ and ground truth class labels.
They include a set of positive-label constraints $P=\{(v_i,y_m): \: v_i \in y_m\}$, indicating that the true class label of $v_i$ is $y_m$, and a set of negative-label constraints $N=\{(v_i,y_m): \: v_i \notin y_m\}$, indicating that the true class label of $v_i$ is not $y_m$.
To form a uniform representation of constraints,  we convert label constraints into pairwise constraints which are more compatible for structural entropy.
For two vertices $v_i$ and $v_j$, the conversion rules are set as follows:
(1) If they both have positive constraints with the same label, an edge exists in $E'$ with a positive value $\gamma_M$ added to the edge weight $\mathbf{W}'_{ij}$.
(2) If they both have positive constraints with different labels, an edge exists in $E'$ with a negative value $\gamma_C$ added to the edge weight $\mathbf{W}'_{ij}$.
(3) If they have positive constraint and negative constraints respectively with the same label, an edge exists in $E'$ with a negative value $\gamma_C$ added to the edge weight $\mathbf{W}'_{ij}$.
% Furthermore, we follow the strategy of Bai \textit{et al.} \cite{bai2020semi} to sufficiently exploit the label constraints.

The constraints are stored in the relation graph $G'$ after construction, where a positive value indicates a must-link relationship and a negative value indicates a cannot-link relationship.
However, this relation graph can be further improved by exploiting \textit{constraint transitivity} and \textit{entailment} \cite{van2017cobra}.
We apply them sequentially on $G'$ after constructing it.

% \subsection{Two-Dimensional SSE.}\label{sec:2dsse}
\subsection{2-D SSE.}\label{sec:2dsse}
In this subsection, we present 2-d SSE modified from 2-d structural entropy  to perform semi-supervised partitioning clustering.
For a graph $G$ associated with a data set $\mathcal{X}$ with different types of constraints, we transform all types of constraints into a uniform formulation and store them in a relation graph $G'$.
We aim to find a graph partitioning $\mathcal{P}$ of $G$ that minimizes the structural entropy of $G$ while also minimizing the number of violated constraints in the meantime.
The optimization objective of two-dimensional structural entropy is defined as follows:
\begin{equation}\label{eq:lp}
    \mathcal{L}^\mathcal{P}(G,G') = \mathcal{H}^\mathcal{P}(G) + \phi \mathcal{E}^\mathcal{P}(G,G'),
\end{equation}
where $\mathcal{E}^\mathcal{P}(G,G')$ is a penalty term for constraints violation, and it is defined as:
\begin{equation}
\begin{aligned}
    % E^\mathcal{P}(G,A) = - \sum_{X \in \mathcal{P}} \sum_{v_i \in X} \frac{g'_i}{vol(G)}\log_2\frac{d_i}{\mathcal{V}_X} \\ 
    % - \sum_{X \in \mathcal{P}} \frac{g'_X}{vol(G)} \log_2\frac{\mathcal{V}_X}{vol(G)},
    \mathcal{E}^\mathcal{P}(G,G') = - \sum_{X \in \mathcal{P}} \frac{g'_X}{\mathcal{V}_G} \log_2\frac{\mathcal{V}_X}{\mathcal{V}_G},
\end{aligned}
\end{equation}
where $g'_X$ is the weight sum of edges in $G'$ between vertices in and not in module $X$, and other notations share the same meaning with notations in Eq. (\ref{eq:hp}).

% The intuition of the penalty term is that we modify $g_X$, i.e., the cut of module $X$ in Eq. \ref{eq:hp} according to the constraints, which is increased when must-link constraints related to $X$ are violated, and decreased when cannot-link constraints related to $X$ are satisfied.
The intuition of the penalty term is that we modify $g_X$, i.e., the cut of module $X$ in Eq. (\ref{eq:hp}) according to the constraints, which is increased when must-link constraints are violated, and decreased when cannot-link constraints are satisfied. 
% When must-link constraints exist and some constraints related to vertices in module $X$ are violated, $g'_X > 0$ and the penalty term $\mathcal{E}^\mathcal{P}(G,G') > 0$, and it is omitted only when two vertices in the must-link vertex pair belong to the same module.
% When cannot-link constraints exist and some constraints related to vertices in module $X$ are satisfied, $g'_X < 0$ and the penalty term $\mathcal{E}^\mathcal{P}(G,G') < 0$ which rewards the satisfaction of constraints, and this reward is omitted only when no cannot-link constraint is satisfied.
% A positive value of $\mathbf{W}'_{ij}$ in $G'$ means $v_i$ and $v_j$ should belong to the same module and a penalty will be added if it is violated, while a negative value of $\mathbf{W}'_{ij}$ means $v_i$ and $v_j$ should belong to different modules and a reward will be added if it is satisfied.
A positive value of $\mathbf{W}'_{ij}$ in $G'$ means $v_i$ and $v_j$ should belong to the same module, and $\mathcal{E}^\mathcal{P} > 0$ if they are not, leading to a penalty added to $\mathcal{L}^\mathcal{P}$.
A negative value of $\mathbf{W}'_{ij}$ in $G'$ means $v_i$ and $v_j$ should belong to different modules, and $\mathcal{E}^\mathcal{P} < 0$ if they are, leading to a reward added to $\mathcal{L}^\mathcal{P}$.
When no constraint exists, i.e., $\mathcal{E}^\mathcal{P} = 0$, we only minimize unsupervised 2-d structural entropy.
In all, $\mathcal{E}^\mathcal{P}$ penalizes modules that violate must-link constraints and rewards modules that satisfy cannot-link constraints.

\noindent\textbf{Minimizing 2-D SSE.}
We minimize 2-d SSE via two operators \textit{merging} \cite{li2016structural} and \textit{moving} on the encoding tree $\mathcal{T}$.
For two sister nodes $\alpha, \beta \in \mathcal{T}$ with associated vertex subsets $X$ and $Y$, node \textit{merging} is defined as: (1) set $X = X \cup Y$, (2) delete $\beta$.
The decrease amount of $\mathcal{L}^\mathcal{P}(G,G')$ is given by:
\begin{equation}
\label{eq:merging}
\begin{split}
    \Delta \mathcal{L}^\mathcal{M}_{X,Y} = 
    \frac{1}{\mathcal{V}_G} [\left(\mathcal{V}_X - g_X - g'_X\right) \log_2  \mathcal{V}_X \\
    + \left(\mathcal{V}_Y - g_Y - g'_Y\right) \log_2  \mathcal{V}_Y \\
    - \left( \mathcal{V}_{X \cup Y} - g_{X \cup Y} - g'_{X \cup Y} \right) \log_2  \mathcal{V}_{X \cup Y} \\
    + \left( g_{X} + g_{Y} - g_{X \cup Y} + g'_X + g'_Y - g'_{X \cup Y} \right)\log_2  \mathcal{V}_G],
\end{split}
\end{equation}
where $\mathcal{M}$ denotes $\mathcal{M}$erging operator, $\mathcal{V}_X$ is the volume of $X$ in $G$, $\mathcal{V}_G$ is the volume of $G$, $g_X$ and $g'_X$ are the cuts of $X$ in $G$ and $G'$, respectively.
For a node $\alpha \in \mathcal{T}$ with associated module $X$ and a vertex $v_i \in X$, the \textit{moving} operator seeks to find a new node $\beta \in \mathcal{T}$ with associated module $Y$ and move $v_i$ from $X$ to $Y$.
The decrease amount of $\mathcal{L}^\mathcal{P}(G,G')$ by removing $v_i$ from $X$ is given by:
\begin{equation}
\label{eq:removing}
\begin{split}
    \Delta \mathcal{L}^\mathcal{R}_{X,v_i} = 
    \frac{\mathcal{V}_X - g_X - g'_X}{\mathcal{V}_G}\log_2\frac{\mathcal{V}_X}{\mathcal{V}_G}\\ 
    - \frac{\mathcal{V}_{X\backslash\{v_i\}} - g_{X\backslash\{v_i\}} -g'_{X\backslash\{v_i\}}}{\mathcal{V}_G} \log_2 \frac{\mathcal{V}_{X\backslash\{v_i\}}}{\mathcal{V}_G},
\end{split}
\end{equation}
where $\mathcal{R}$ denotes vertex $\mathcal{R}$emoving and $X\backslash\{v_i\}$ denotes removing vertex $v_i$ from $X$.
The increase amount of $\mathcal{L}^\mathcal{P}(G,G')$ by inserting $v_i$ into $Y$ is given by:
\begin{equation}
\label{eq:inserting}
\begin{split}
    \Delta \mathcal{L}^\mathcal{I}_{Y,v_i} = 
    - \frac{\mathcal{V}_Y - g_Y - g'_Y}{\mathcal{V}_G} \log_2 \frac{\mathcal{V}_Y}{\mathcal{V}_G} \\
    + \frac{\mathcal{V}_{Y \cup \{v_i\}} - g_{Y \cup \{v_i\}} - g'_{Y \cup \{v_i\}}}{\mathcal{V}_G} \log_2 \frac{\mathcal{V}_{Y \cup \{v_i\}}}{\mathcal{V}_G},
\end{split}
\end{equation}
where $\mathcal{I}$ denotes vertex $\mathcal{I}$nserting and $Y \cup \{v_i\}$ denotes inserting $v_i$ into $Y$.
We initialize $\mathcal{T}$ to contain a root node $\lambda$ and $n$ leaves where each leaf is associated with one vertex in $G$, and then sequentially apply \textit{merging} and \textit{moving} operators until convergence.
The optimization procedure is summarized in Algorithm \ref{algo:2dsse}.

\renewcommand{\algorithmicrequire}{\textbf{Input:}}
\renewcommand{\algorithmicensure}{\textbf{Output:}}
\begin{algorithm}[t]
\caption{2-d SSE minimization}
\label{algo:2dsse}
\begin{algorithmic}[1]
    \REQUIRE $G=(V,E,\mathbf{W})$, $G'=(V,E',\mathbf{W}')$
    \ENSURE Encoding tree $\mathcal{T}$ and partitioning $\mathcal{P}$
    \STATE Initialize $\mathcal{T}$ containing all vertices as tree leaves
    \STATE // \textit{Merging} stage
    \REPEAT
        \STATE Merge a chosen module pair $(X,Y)$ into $X \cup Y$ condition on $\arg\max_{X,Y} \{ \Delta \mathcal{L}^\mathcal{M}_{X,Y} \}$ via Eq. (\ref{eq:merging})
        \STATE Update $\Delta \mathcal{L}^\mathcal{M}$ for module pairs connected to $X$ or $Y$
    \UNTIL $\Delta \mathcal{L}^\mathcal{M} < 0$ for all module pairs
    \STATE // \textit{Moving} stage
    \REPEAT
        \FOR{each vertex $v_i \in V$}
        \STATE Remove vertex $v_i$ from the original module $X$
        \STATE Insert node $v_i$ into a chosen module $Y$ condition on $\arg\max_{Y} \{ \Delta \mathcal{L}^\mathcal{R}_{X,v_i} - \Delta \mathcal{L}^\mathcal{I}_{Y,v_i} \}$ via Eqs. (\ref{eq:removing}) and (\ref{eq:inserting})
        \ENDFOR
    \UNTIL $\mathcal{L}^\mathcal{P}(G,G')$ converges
\end{algorithmic}
\end{algorithm}

In both \textit{merging} stage and \textit{moving} stage, $\mathcal{L}^\mathcal{P}$ decreases after every iteration, and it converges when no improvement can be made.
The time complexity of \textit{merging} stage is $O(n{\log^2n})$ \cite{li2016structural}.
In the \textit{moving} stage, each iteration requires calculating $\Delta \mathcal{L}^\mathcal{R}_{X,v_i}$ and $\Delta \mathcal{L}^\mathcal{I}_{Y,v_i}$ for every vertex $v_i$ and every possible module $Y$, which takes the time of $O(nl)$.
Taken together, the time complexity of Algorithm \ref{algo:2dsse} is $O(n\log^2n+nlt)$, where $n$, $l$ and $t$ denote the number of vertices, modules, and iterations respectively.

\subsection{High-D SSE.}
% In this subsection, we extend two-dimensional SSE into high-dimensional SSE modified from high-dimensional structural entropy to perform semi-supervised hierarchical clustering.
Hereafter, we generalize 2-d SSE into high-d SSE to perform semi-supervised hierarchical clustering.
For a graph $G$ associated with data set $\mathcal{X}$ and constraints stored in a relation graph $G'$, we aim to find an encoding tree with the height of $K>2$ to form a semi-supervised hierarchical clustering of vertices in $G$.
Following the definition of 2-d SSE in Section \ref{sec:2dsse}, we define the optimization objective of high-d SSE as follows:
\begin{equation}
    \mathcal{L}^\mathcal{T}(G,G') = \mathcal{H}^\mathcal{T}(G) + \phi \mathcal{E}^\mathcal{T}(G,G'),
\end{equation}
where $\mathcal{E}^\mathcal{T}(G,G')$ is a penalty term for constraints violation, and it is defined as:
\begin{equation}
    \mathcal{E}^\mathcal{T}(G,G') = \sum_{\alpha \in \mathcal{T}, 1<|T(\alpha)|<|V|} - \frac{g'_\alpha}{\mathcal{V}_G}\log_2\frac{\mathcal{V}_\alpha}{\mathcal{V}_{\alpha^-}},
\end{equation}
where $g'_\alpha$ is the cut of $\alpha$ in $G'$, $|T(\alpha)|$ is the number of vertices in subset $T(\alpha)$ associated to $\alpha$, and other notations share the same meaning with notations in Eq. (\ref{eq:ht}).
For each node except for leaves in $\mathcal{T}$, the penalty term penalizes the violation of must-link constraints and rewards the satisfaction of cannot-link constraints.
% where $g'_\alpha$ is the weight sum of edges in $A$ between vertices in and not in the vertex set of $\alpha$, i.e., $\mathcal{T}_\alpha$, and other notations share the same meaning with notations in Eq. \ref{eq:ht}.
% Through changes in the cut of tree node $\alpha$, penalties are added to $L^\mathcal{T}$ if constraints are violated and rewards are added to $L^\mathcal{T}$ if constraints are satisfied.

\noindent\textbf{Minimizing High-D SSE}.
We minimize high-d SSE via two operators \textit{stretching} and \textit{compressing} on the encoding tree $\mathcal{T}$ \cite{pan2021information}.
For a pair of sister nodes $(\alpha, \beta ) \in \mathcal{T}$ whose parent is $\gamma$, node \textit{stretching} is defined as inserting a new node $\delta$ between $\gamma$ and $\left(\alpha, \beta \right)$, i.e., (1) set $\alpha^- = \delta$, (2) set $\beta^- = \delta$, (3) set $\delta^- = \gamma$.
The decrease amount of $\mathcal{L}^\mathcal{T}(G,G')$ is given by:
\begin{equation}
\label{eq:stretching}
\begin{split}
    \Delta \mathcal{L}^\mathcal{S}_{\alpha,\beta} = 
    \frac{g_\alpha + g_\beta - g_\delta + g'_\alpha + g'_\beta - g'_\delta}{\mathcal{V}_G} \log_2 \frac{\mathcal{V}_\gamma}{\mathcal{V}_\delta},
\end{split}
\end{equation}
where $\mathcal{S}$ denotes node $\mathcal{S}$tretching.
Applying \textit{stretching} on the initial encoding tree iteratively results in a binary encoding tree $\mathcal{T}_b$.
For a node $\alpha \in \mathcal{T}$ contains a set of children $\{ \beta_1,...,\beta_m \}$ and its parent is $\gamma$, node \textit{compressing} is defined as: (1) remove node $\alpha$, (2) for each child node $\beta_i$ of $\alpha$, set $\beta_i^- = \gamma$.
The decrease amount of $\mathcal{L}^\mathcal{T}(G,G')$ is given by:
\begin{equation}
\label{eq:compressing}
\begin{split}
    \Delta \mathcal{L}^\mathcal{C}_{\alpha} = 
    \frac{\sum\limits_{i} g_{\beta_i} + \sum\limits_{|T(\beta_i)|>1} g'_{\beta_i} - g_\alpha - g'_\alpha}{\mathcal{V}_G} \log_2 \frac{\mathcal{V}_\alpha}{\mathcal{V}_\gamma},
\end{split}
\end{equation}
where $\mathcal{C}$ denotes node $\mathcal{C}$ompressing.
Applying \textit{compressing} on the binary encoding tree results in a multinary encoding tree.
By restricting the height of the encoding tree to be less than the required height $K$, we can obtain the $K$-d encoding tree.
We summarize this optimization procedure in Algorithm \ref{algo:hdsse}.
For a graph $G$ with $n$ vertices and $m$ edges, the time complexity of Algorithm \ref{algo:hdsse} is $O(h_{max}(mlogn+n))$, where $h_{max}$ is the height of $\mathcal{T}_b$.

\renewcommand{\algorithmicrequire}{\textbf{Input:}}
\renewcommand{\algorithmicensure}{\textbf{Output:}}
\begin{algorithm}[t]
\caption{High-d SSE minimization}
\label{algo:hdsse}
\begin{algorithmic}[1]
    \REQUIRE $G=(V,E,\mathbf{W})$, $G'=(V,E',\mathbf{W}')$, height $K$
    \ENSURE Binary tree $\mathcal{T}_b$ and height $K$ tree $\mathcal{T}_K$
    \STATE Initialize $\mathcal{T}$ with a root node $\lambda$ and all vertices as tree leaves
    \STATE // \textit{Stretching} stage
    \REPEAT
    \STATE Stretch a chosen node pair $\{ \alpha, \beta \}$ condition on $\arg\max_{\alpha, \beta} \{ \Delta \mathcal{L}^\mathcal{S}_{\alpha, \beta}\}$ via Eq. (\ref{eq:stretching})
    \STATE Update $\Delta \mathcal{L^\mathcal{S}}$ for node pairs connected to $\alpha$ or $\beta$
    \UNTIL The children number of $\lambda$ is two, resulting in binary tree $\mathcal{T}_b$
    \STATE // \textit{Compressing} stage
    \REPEAT
    \STATE Remove a chosen tree node $\alpha \in \mathcal{T}$ condition on $\arg\max_{\alpha} \{ \Delta \mathcal{L}^\mathcal{C}_\alpha \}$ via Eq. (\ref{eq:compressing})
    \UNTIL Height of encoding tree $\mathcal{T}$ is not larger than $K$, resulting in $\mathcal{T}_K$
\end{algorithmic}
\end{algorithm}

% Convergence analysis and time complexity analysis.
\section{Experiments}\label{sec:experiments}
% Our proposed SSE method is capable of performing both semi-supervised partitional clustering and hierarchical clustering.
Our proposed SSE method is capable of tackling both semi-supervised partitioning clustering and hierarchical clustering.
% To evaluate SSE in terms of these two tasks, we design two groups of experiments comparing SSE with semi-supervised partitional clustering baselines (Section \ref{sec:par}) and semi-supervised hierarchical clustering baselins (Section \ref{sec:hie}).
Regarding the evaluation for both tasks, we design two groups of experiments, in which we compare SSE against established baselines for semi-supervised partitioning clustering (Section \ref{sec:par}) and semi-supervised hierarchical clustering (Section \ref{sec:hie}).

\subsection{Semi-Supervised Partitioning Clustering.}
\label{sec:par}

\begin{table*}[t]
\renewcommand{\arraystretch}{0.95}
% \caption{Performance of SSE for partitional clustering on machine learning datasets. \textbf{Bold}: the best performance on each group of methods.}
\caption{Performance of comparison methods for partitioning clustering on five clustering datasets. \textbf{Bold}: the best performance on each group of methods.}
\label{tab:pml}
\centering
\begin{tabular}{l|l|cc|cc|cc|cc|cc}
\toprule
            & \multicolumn{1}{l|}{\multirow{2}{*}{Method\%}} & \multicolumn{2}{c|}{Yale} & \multicolumn{2}{c|}{ORL}  & \multicolumn{2}{c|}{COIL20} & \multicolumn{2}{c|}{Isolet}& \multicolumn{2}{c}{OpticalDigits} \\ \cline{3-12}
             & \multicolumn{1}{l|}{} & ARI$\uparrow$ & NMI$\uparrow$ & ARI$\uparrow$ & NMI$\uparrow$ & ARI$\uparrow$ & NMI$\uparrow$ & ARI$\uparrow$ & NMI$\uparrow$ & ARI$\uparrow$ & NMI$\uparrow$ \\
\midrule
                & SE & 28.12 & 54.78 & 59.15 & 85.31  & 68.22 & 86.34 & 56.01 & 83.14 & 69.89 & 79.69 \\
\midrule
\multirow{5}{*}{\rotatebox{90}{\makecell[c]{Pairwise}}}  
               & PCPSNMF & 24.80 & 54.11 & 52.02 & 81.86 & 51.49 & 80.75 & 38.39 & 69.15 & 48.82 & 68.71  \\
               & OneStepPCP & 25.50 & 52.22 & 40.58 & 78.14 & 52.81 & 79.70 & 49.93 & 76.01 & 77.90 & 86.02  \\
               & CMS & 07.06 & 35.54 & 29.37 & 73.18 & 59.81 & 78.32 & 48.77 & 77.38 & \textbf{88.75} & \textbf{91.17} \\
               & SC-MPI & 32.76 & 59.82 & 49.28 & 82.29 & 59.89 & 82.83 & 47.16 & 72.75 & 52.39 & 71.35 \\
               & SSE (Ours)  & \textbf{37.12} & \textbf{61.37} & \textbf{65.42} & \textbf{87.51} & \textbf{75.36} & \textbf{87.50} & \textbf{61.37} & \textbf{82.77} & 77.57 & 84.34 \\ 
\midrule
\multirow{5}{*}{\rotatebox{90}{\makecell[c]{Label}}}
                & Seeded-KMeans & 25.21 & 52.06 & 46.35 & 78.56 & 67.59 & 81.40 & \textbf{66.48} & 81.13 & 73.52 & 77.89 \\
                & S4NMF & 23.85 & 49.60 & 47.23 & 77.08 & 62.48 & 79.34 & 57.05 & 77.32 & 84.72 & 88.32 \\
                & LpCNMF & 13.35 & 39.67 & 32.55 & 70.01 & 74.72 & 88.78 & 59.24 & 81.89 & 90.77 & \textbf{93.80} \\
                & SC-MPI &   20.91 & 50.82 & 26.28 & 70.73 & \textbf{89.18} & \textbf{94.21} & 64.43 & 80.38 & \textbf{93.04} & 93.41 \\
                & SSE (Ours) & \textbf{33.48} & \textbf{58.62} & \textbf{61.26} & \textbf{86.01} & 75.10 & 87.63 & 58.62 & \textbf{83.13} & 76.60 & 84.12  \\
\bottomrule
\end{tabular}
\end{table*}

\begin{table*}[t]
\renewcommand{\arraystretch}{0.95}
\caption{Performance of comparison methods for partitioning clustering on four RNA-seq datasets. \textbf{Bold}: the best performance on each group of methods.}
\label{tab:pbio}
\centering
\begin{tabular}{l|l|cc|cc|cc|cc}
\toprule
            & \multicolumn{1}{l|}{\multirow{2}{*}{Method\%}} & \multicolumn{2}{c|}{10X PBMC} & \multicolumn{2}{c|}{Mouse bladder} & \multicolumn{2}{c|}{Worm neuron} & \multicolumn{2}{c}{Human kidney}  \\ \cline{3-10}
             & \multicolumn{1}{l|}{} & ARI$\uparrow$ & NMI$\uparrow$ & ARI$\uparrow$ & NMI$\uparrow$ & ARI$\uparrow$ & NMI$\uparrow$ & ARI$\uparrow$ & NMI$\uparrow$  \\
\midrule
                & SE  & 63.89 & 74.80 & 67.41 & 77.13 & 20.90 & 41.70 & 54.20 & 72.86  \\
\midrule
\multirow{5}{*}{\rotatebox{90}{\makecell[c]{Pairwise}}}  
               & PCPSNMF & 16.41 & 29.68 & 13.55 & 40.02 & 09.45 & 21.48 & 13.46 & 28.66  \\
               & OneStepPCP & 43.08 & 58.29 & 44.51 & 64.86 & 19.39 & 44.42 & 40.21 & 55.26  \\
               & CMS  & 08.58 & 10.40 & 08.28 & 10.56 & 00.27 & 01.40 & 07.18 & 12.26 \\
               & SC-MPI & 20.24 & 30.57 & 18.70 & 40.88 & 08.98 & 17.03 & 18.12 & 31.78 \\
               & SSE (Ours) & \textbf{74.87} & \textbf{76.84} & \textbf{62.33} & \textbf{74.70} & \textbf{22.17} & \textbf{45.05} & \textbf{62.59} & \textbf{75.81}  \\ 
\midrule
\multirow{5}{*}{\rotatebox{90}{\makecell[c]{Label}}}
                & Seeded-KMeans & 67.56 & 71.07 & 38.40 & 62.63 & 07.07 & 34.24 & 17.19 & 39.74  \\
                & S4NMF & 18.28 & 28.20 & 26.27 & 43.79 & 08.90 & 15.54 & 24.33 & 39.07 \\
                & LpCNMF &  44.40 & 62.94 & 44.64 & 73.02 & 34.67 & \textbf{56.37} & 45.90 & 64.88 \\
                & SC-MPI &   48.28 & 63.34 & 38.65 & 55.23 & \textbf{37.82} & 46.58 & 56.93 & 60.36 \\
                & SSE (Ours) &  \textbf{74.15} & \textbf{77.05} & \textbf{63.11} & \textbf{75.38} & 28.43 & 46.94 & \textbf{65.45} & \textbf{78.23} \\
\bottomrule
\end{tabular}
\end{table*}

\begin{table*}[t]
\renewcommand{\arraystretch}{0.95}
\setlength{\tabcolsep}{4pt}
\caption{Performance of comparison methods for semi-supervised hierarchical clustering. \textbf{Bold}: the best performance on each group of methods.}
\label{tab:hie}
\centering
\begin{tabular}{l|ccc|ccc|ccc|ccc}
\toprule
\multicolumn{1}{l|}{\multirow{2}{*}{Method\%}} & \multicolumn{3}{c|}{Wine} & \multicolumn{3}{c|}{Heart} & \multicolumn{3}{c|}{Br. Cancer} & \multicolumn{3}{c}{Australian} \\ \cline{2-13}
\multicolumn{1}{l|}{} & DP$\uparrow$ & ARI$\uparrow$ & NMI$\uparrow$ & DP$\uparrow$ & ARI$\uparrow$ & NMI$\uparrow$ & DP$\uparrow$ & ARI$\uparrow$ & NMI$\uparrow$ & DP$\uparrow$ & ARI$\uparrow$ & NMI$\uparrow$ \\ \midrule
SE          & 84.87 & 73.85 & 74.19 & 61.97 & 07.70 & 21.30 & 95.75 & 88.00 & 80.93 & 57.02 & 01.95 & 08.03  \\
SpecWRSC   & 84.87 & 76.94 & 77.10 & 70.13 & \textbf{33.13} & \textbf{29.23} & 95.68 & \textbf{88.55} & \textbf{81.57} & 54.92 & -00.78 & 01.21 \\ 
\midrule
COBRA       & 86.50 & 81.26 & 80.07 & 64.12 & 26.07 & 20.92 & 92.23 & 82.38 & 72.41 & 66.55 & 32.03 & 24.61 \\
SemiMulticons  & 90.52 & 82.99 & 82.69 & 69.29 & 28.44 & 26.91 & 92.68 & 82.77 & 73.79 & 72.13 & \textbf{39.47} & \textbf{33.83}  \\
SSE (Ours)    & \textbf{92.88} & \textbf{85.27} & \textbf{83.61} & \textbf{71.90} & 28.08 & 26.36 & \textbf{96.53} & 82.88 & 76.08 & \textbf{74.52} & 34.19 & 28.17 \\ \bottomrule
% Performance can still be improved.
\end{tabular}
\end{table*}

In this part, we aim to evaluate the performance of SSE on semi-supervised partitioning clustering.
We conduct experiments on five clustering datasets including face image data (Yale and ORL), object image data (COIL20), spoken letter recognition data (Isolet), and handwritten digit data (OpticalDigits) following Bai \textit{et al.} \cite{bai2020semi}, whose size ranges from 165 to 5620.
We also conduct experiments on four single-cell RAN-seq datasets including 10X PBMC, Mouse bladder, Worm neuron, and Human kidney taken from Tian \textit{et al.} \cite{tian2021model}.
We choose the data preprocessed by the original authors to contain 2100 randomly sampled cells on the top 2000 highly dispersed genes in each dataset.
% Details of these datasets are summarized in Table \ref{tab:datasets}.
We adopt two metrics including Adjusted Rand Index (ARI) \cite{hubert1985comparing} and Normalized Mutual Information (NMI) \cite{strehl2002cluster} for partitioning clustering performance evaluation.
All experiments are repeated 10 times.

\noindent\textbf{Baselines.}
We compare SSE with a variety of baseline methods, including an unsupervised clustering method based on structural entropy minimization, three semi-supervised clustering methods with pairwise constraints, three semi-supervised clustering methods with label constraints, and one semi-supervised clustering method with both pairwise constraints and label constraints.
The unsupervised clustering based on structural entropy minimization is optimized by the \textit{merging} operator (SE \cite{li2016structural}).
For semi-supervised clustering methods with pairwise constraints, we consider pairwise constraint propagation-induced symmetric NMF (PCPSNMF \cite{wu2018pairwise}), jointly optimized pairwise constraint propagation and spectral clustering (OneStepPCP \cite{jia2020joint}), and constrained mean shift clustering (CMS \cite{schier2022constrained}).
For semi-supervised clustering methods with label constraints, we consider seeded semi-supervised KMeans (Seeded-KMeans \cite{basu:ml02}), self-supervised semi-supervised NMF (S4NMF \cite{chavoshinejad2023self}), and label propagation based constrained NMF (LpCNMF \cite{liu2023constrained}).
SC-MPI \cite{bai2020semi} is a semi-supervised spectral clustering method capable of dealing with different types of constraints.
Since SSE and SC-MPI are capable of dealing with both pairwise constraints and label constraints, they are compared in both groups.

\noindent\textbf{Implementation Details.}\label{sec:imple1}
We construct graph $G$ from the given dataset $\mathcal{X}$ by calculating the similarity between data points and then sparsify it into a $p$-nearest-neighbor graph by retaining $p$ most significant edges from each node.
For a given $\mathcal{X}$ with $n$ data points divided into $k$ clusters according to the ground truth, we empirically set $p$ to be $\lfloor 20k/ \log^2_2 n \rfloor + 1$, since the number of clusters by minimizing $\mathcal{H}^\mathcal{P}$ is approximately $\Theta(p\log^2_2 n)$ \cite{li2016structural}.
For five clustering datasets, the similarity is defined by a Gaussian kernel with kernel width $\sigma = 10$.
For four single-cell RNA-seq datasets, the similarity is defined as cosine similarity since the features of these datasets are sparse.

We generate constraints using the ground truth class labels from the datasets.
For experiments with pairwise constraints, we set the number of must-link constraints the same as cannot-link constraints to be $0.2n$.
For experiments with label constraints, we set the number of positive constraints the same as negative constraints to be $0.1n$.
The parameters $\gamma_M$ and $\gamma_C$ control the role of constraints, we define them following Bai \textit{et al.} \cite{bai2020semi}.
For a pair of data points $(x_i,x_j)$ with similarity $\mathbf{W}_{ij}$, we define $\gamma_M = max(\mathbf{W}) - \mathbf{W}_{ij}$, where $max(\mathbf{W})$ is the maximum similarity between all data points.
The process of constraints conversion, \textit{constraints transitivity} and \textit{entailment} usually lead to more negative values than positive values in $G'$.
In order to balance them, we define $\gamma_C= \rho( min(\mathbf{W}) - \mathbf{W}_{ij})$, where $\rho$ is the ratio between the number of positive values and negative values in $G'$, $min(\mathbf{W})$ is the minimum similarity between all data points.
The parameter $\phi$ balances the importance between input data and constraints, it is empirically set as $\phi = 2$.

\noindent\textbf{Experimental Results.}
The experimental results on five clustering datasets are presented in Table \ref{tab:pml}.
Three groups of methods, i.e., unsupervised clustering, semi-supervised clustering with pairwise constraints, and semi-supervised clustering with label constraints, are compared separately.
SSE with pairwise constraints outperforms its unsupervised baseline SE on all datasets and outperforms baseline methods in the pairwise constraint group on four out of five datasets.
SSE with label constraints outperforms SE on all datasets and outperforms baseline methods in the label constraint group on three out of five datasets.
% Both SSE with pairwise constraints and SSE with label constraints fail to outperform SE on the COIL20 dataset, they achieve slightly lower performance.
% This indicates that SE without constraints captures the cluster structure of COIL20 well, introducing additional pairwise constraints or label constraints impedes the performance.

The experimental results on four single-cell RNA-seq datasets are presented in Table \ref{tab:pbio}.
SSE with pairwise constraints outperforms SE on three out of four datasets and outperforms baseline methods in the pairwise constraint group on all datasets.
SSE with label constraints outperforms SE on three out of four datasets and outperforms baseline methods in the label constraint group on three out of four datasets.
In all, SSE effectively incorporates prior information in the forms of pairwise constraints and label constraints, and achieves high clustering accuracy on both clustering datasets and single-cell RNA-seq datasets.

\subsection{Semi-Supervised Hierarchical Clustering.}
\label{sec:hie}
In this part, we aim to evaluate the performance of SSE on semi-supervised hierarchical clustering.
% We conduct experiments on four datasets including Wine, Heart, Br. Cancer, and Australian following Chierchia and Perret \cite{chierchia2019ultrametric}.
We conduct experiments on four datasets downloaded from the LIBSVM webpage \footnote{https://www.csie.ntu.edu.tw/$\sim$cjlin/libsvmtools/datasets/} following Chierchia and Perret \cite{chierchia2019ultrametric}, whose size ranges from 175 to 690.
We adopt three metrics including Dendrogram Purity (DP) \cite{heller2005bayesian,yadav2019supervised}, ARI \cite{hubert1985comparing}, and NMI \cite{strehl2002cluster} for hierarchical clustering performance evaluation.
DP is a holistic measure of a cluster tree, which is defined as the weighted average purity of each node of the tree with respect to a ground truth labelization of the tree leaves.
We take the cluster tree of SSE from the binary encoding tree $\mathcal{T}_b$.
ARI and NMI require partitioning clustering results from the cluster trees.
We perform the $compressing$ operator until the height of the encoding tree is two to obtain the partitioning clustering results.
% may be other options?
For other methods, we choose the largest tree nodes from the cluster tree as the partitioning clustering results.
All experiments are repeated 10 times.

\begin{figure*}[t]
\centering
\includegraphics[width=1\linewidth]{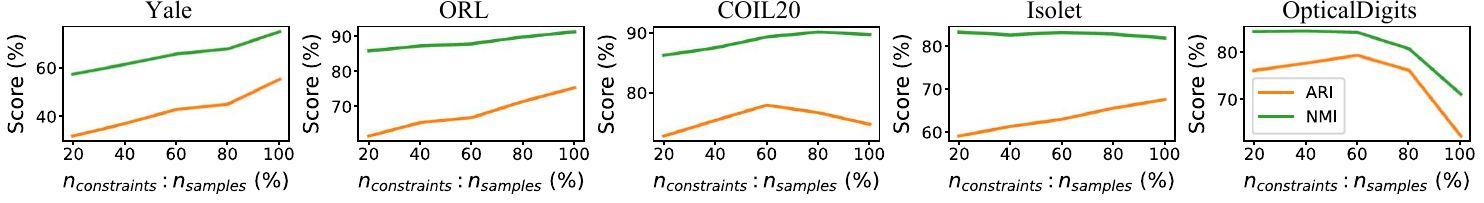}
\vspace{-0.5cm}
% \captionsetup{width=0.8\textwidth}
\caption{Performance of SSE for semi-supervised partitioning clustering with different constraint amounts.}
\label{fig:sens_pair}
\end{figure*}

% \begin{figure*}[t]
% \centering
% \includegraphics[width=1\linewidth]{figures/sensitivity_label.pdf}
% \vspace{-0.8cm}
% % \captionsetup{width=0.8\textwidth}
% \caption{Label constraints.}
% \label{fig:sens_label}
% \end{figure*}

\begin{figure*}[t]
\centering
\includegraphics[width=0.8\linewidth]{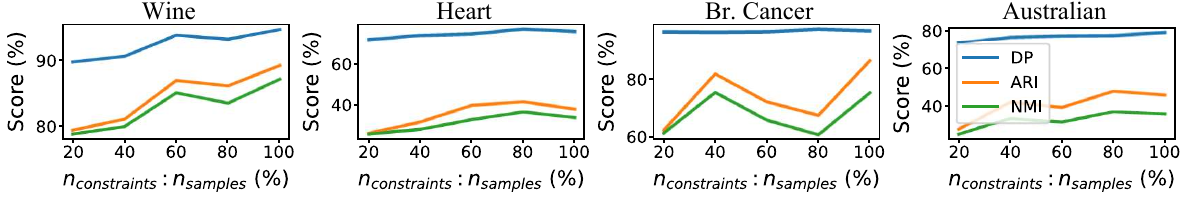}
\vspace{-0.13cm}
% \captionsetup{width=0.8\textwidth}
\caption{Performance of SSE for semi-supervised hierarchical clustering with different constraint amounts.}
\label{fig:sens_hierar}
\end{figure*}

\noindent\textbf{Baselines.}
We compare SSE with two unsupervised hierarchical clustering methods and two semi-supervised hierarchical clustering methods.
For unsupervised hierarchical clustering methods, we consider structural entropy minimized by \textit{stretching} operator and \textit{compressing} operator (SE \cite{pan2021information}) and sublinear time graph-based hierarchical clustering (SpecWRSC \cite{kapralov2023learning}).
For semi-supervised hierarchical clustering methods, we consider merging-based active clustering (COBRA \cite{van2017cobra}) and closed pattern mining based semi-supervised consensus clustering (SemiMulticons \cite{yang2022semi}).

\noindent\textbf{Implementation Details.}
We construct graph $G$ from the given dataset $\mathcal{X}$ by calculating cosine similarity between data points and then sparsify it into a 5-nearest-neighbor graph by retaining 5 significant edges from each node.
We generate $0.2n$ must-link constraints and $0.2n$ cannot-link constraints randomly for all methods except for COBRA, for which we generate $0.2n$ positive constraints due to its requirements.
We set $\gamma_M= max(\mathbf{W}) \mathbf{W}_{ij}$ and $\gamma_C=\rho ( min(\mathbf{W}) \mathbf{W}_{ij} )$.
% , where $\mathbf{W}_{ij}$ is the similarity between data points $x_i$ and $x_j$, $max(\mathbf{W})$ and $min(\mathbf{W})$ are maximum and minimum similarity between all data points, and $\delta$ is the ratio between the number of positive values and negative values in $G'$.
The parameter $\phi$ is set to be 2.

\noindent\textbf{Experimental Results.}
The experimental results of semi-supervised hierarchical clustering are presented in Table \ref{tab:hie}.
% SSE achieves the highest DP value on all datasets and achieves comparable values of ARI and NMI with baselines.
% In addition, SSE outperforms SE with higher values of DP on all datasets and higher values of ARI and NMI on three out of four datasets.
SSE achieves the highest DP values and outperforms SE in terms of DP on all datasets, indicating that the cluster trees of SSE have the highest holistic quality.
The ARI and NMI values of SSE are comparable with baselines and higher than SE on three out of four datasets.
In all, SSE achieves high clustering accuracy on semi-supervised hierarchical clustering.

\subsection{Sensitivity Analysis.}
The number of constraints has a great impact on the performance of semi-supervised clustering and a larger number of constraints is usually thought to lead to better performance.
We evaluate the performance of SSE for partitioning clustering with different amounts of pairwise constraint, as shown in Figure \ref{fig:sens_pair}.
The ARI and NMI values are generally larger with more constraints except for OpticalDigits.
For this dataset, SSE makes too many clusters when the constraints are more than $0.6n$, leading to poor performance.
The cause of this problem is that the \textit{merging} stage in Algorithm \ref{algo:2dsse} stops earlier than expected, which calls for a better optimization algorithm.
We also evaluate the performance of SSE for hierarchical clustering with different amounts of pairwise constraint, as shown in Figure \ref{fig:sens_hierar}.
The growth of DP values can be barely seen, since the DP values of all amounts of constraint are very high.
The ARI and NMI values grow a lot with more constraints provided.
In all, SSE performs better with more constraints under most circumstances.

% \subsection{Efficiency}
\section{Related Work}
% \subsection{Semi-supervised clustering}
Semi-supervised clustering methods incorporate prior information into the process of clustering to enhance clustering quality and better align user preferences, and have attracted great interest in recent years.
Prior information can take different forms of constraints, among them pairwise constraints and label constraints are mostly used.
Pairwise constraints indicate whether a pair of data points should be in the same cluster or not \cite{nie2021semi}.
Many methods that incorporate pairwise constraints have been proposed, such as semi-supervised spectral clustering \cite{bai2020semi}, semi-supervised NMF clustering \cite{wu2018pairwise}, and semi-supervised density peak clustering \cite{schier2022constrained}.
Label constraints reveal class labels of some data points, specifying whether they belong to certain classes or not.
These constraints can be used through label propagation \cite{liu2023constrained} or penalizing violated data points\cite{chavoshinejad2023self}.

% , or initializing the seed for seeded-KMeans \cite{basu:ml02}.

\section{Conclusion}
In this paper, we propose SSE, a novel and more general semi-supervised clustering method that can integrate different types of constraints. 
We give a uniform formulation of pairwise constraints and label constraints and make them both compatible with SSE. 
Moreover, SSE can perform both semi-supervised partitioning clustering and hierarchical clustering, thanks to the structural entropy measure that it is based on. 
We conduct extensive experiments on nine clustering datasets and compare SSE with eleven baselines, justifying the superiority of SSE on high clustering accuracy. 
We also apply SSE to four single-cell RNA-seq datasets for cell clustering, demonstrating its functionality in biological data analysis. 
Future work on SSE may focus on better optimization algorithms.

% In this paper, we have presented SSE, a novel and general semi-supervised clustering method.
% Facing the constraints of different types, we introduce a uniform representation of pairwise constraints and label constraints and make them both applicable to SSE.
% Furthermore, thanks to the advantage of structural entropy which SSE is based on, SSE is capable of performing both semi-supervised partitioning clustering and hierarchical clustering, resulting in higher generalization ability compared to other methods.
% Our extensive experiments on nine clustering datasets against eleven baselines confirm the high clustering accuracy of SSE.
% Moreover, experiments on four single-cell RNA-seq datasets for cell clustering demonstrate the functionality of SSE in biological data analysis.
% Future work on SSE might focus on better optimization algorithms.

\section*{Acknowledgments}
The corresponding author is Hao Peng. 
This work is supported by National Key R\&D Program of China through grant 2021YFB1714800, NSFC through grants 61932002 and 62322202, Beijing Natural Science Foundation through grant 4222030, CCF-DiDi GAIA Collaborative Research Funds for Young Scholars, and the Fundamental Research Funds for the Central Universities.

\bibliographystyle{siamplain}
\bibliography{references}

\end{document}